\documentclass[10pt]{article}
\usepackage[margin=1in]{geometry}
\usepackage{amsmath,amssymb}
\usepackage{booktabs}
\usepackage{graphicx}
\graphicspath{{./}}
\usepackage{xcolor}
\usepackage[numbers,sort&compress]{natbib}
\usepackage{hyperref}

\newcommand{\rinse}{\textsc{rinse}}

\title{Smoothness Ranks Skill, Not Success:\\
An Audit of Published Trajectory-Smoothness Curation Metrics\\
Under Episode-Length Controls\thanks{Code and scored data:
\url{https://github.com/imjbassi/smoothness-metric-audit} (released on publication).}}

\author{Jaiveer Bassi \\ Independent Researcher \\ \texttt{jaiveerbassi@yahoo.com}}
\date{}

\begin{document}
\maketitle

\begin{abstract}
Trajectory smoothness has been proposed as a cheap, training-free signal for
curating imitation-learning demonstrations, most recently by \rinse{}, which
introduces Spectral Arc Length (SAL) and a contact-aware Trajectory-Envelope
Distance (TED) and reports downstream gains on robomimic. Recent audit work
showed that generic curation metrics on a synthetic-defect LIBERO testbed
largely exploit episode length rather than demonstration content. We test
whether the \emph{specific published metrics} survive the corresponding
controls on \emph{real} data. On robomimic can/mh, SAL separates
better from worse operators with AUROC 0.952, but episode length alone
achieves 0.957, and inside the tightest duration-matched band SAL falls to
0.628 while raw mean jerk, the baseline \rinse{} argues
against, rises to 0.896. On can/mg, where every episode is exactly 150
steps and length cannot carry information, SAL is at chance (0.454) and TED
is inverted (0.376) against the task-success label, while jerk reaches
0.706. Training GMM behavior-cloning policies on each metric's curated
subset (10 seeds per condition), curating by TED costs roughly a third of
downstream success relative to random selection of the same size (0.240
vs.\ 0.360, $p{=}0.0001$, surviving Bonferroni correction); path length is
directionally harmful with weaker support, while SAL and jerk are not
statistically distinguishable from not curating at all. Replicating the grid
under a diffusion policy leaves TED ranked worst under both architectures,
though at five seeds its deficit there is no longer resolvable, and reverses
jerk from inert to significantly better than both baselines, so the curation
decision does not transfer unchanged across the two policy configurations
we tested.
Inspecting individual trajectories shows why: on this task successful
manipulation requires sustained high-deviation motion through the contact
phase, while failure manifests as low-amplitude hovering that a smoothed
envelope follows closely, so an envelope-based smoothness metric measures
commitment with the sign flipped. The length
confound replicates on a second task: on square/mh, SAL correlates $+0.96$
with episode duration and its skill AUROC falls from 0.898 to 0.480 inside a
duration-matched band, while jerk rises from 0.431 to 0.896. Our results
extend the detection-vs-curation decoupling to named, published metrics, on
operator labels rather than injected defects, and to a benchmark where the
length confound is excluded by construction rather than by truncation.
\end{abstract}

\section{Introduction}

Behavior cloning inherits the quality of its demonstrations, and a growing
literature proposes to score and filter demonstrations automatically.
Training-free trajectory statistics are the cheapest family of such scores,
and the most developed recent proposal is \rinse{}~\citep{rinse}, which ranks
demonstrations by two smoothness metrics: Spectral Arc Length (SAL), a
frequency-domain measure adapted from rehabilitation science, and
Trajectory-Envelope Distance (TED), a contact-aware spatial deviation from a
B\'ezier-smoothed reference. \rinse{} reports that filtering by these metrics
matches or beats full-data training on robomimic tasks.

A recent line of audit work complicates this picture.
Bedi~\citep{bedi-audit} showed on a controlled testbed that action-only
curation scorers are blind or inverted on structural defects, and the
follow-up~\citep{bedi-policy} showed that five of seven curation metrics on a
LIBERO testbed achieved near-perfect detection AUROC by measuring episode
length rather than demonstration content, and that detection accuracy and
curation value are sharply decoupled. Those results, however, were obtained
with \emph{generic} metric implementations (a SPARC-style smoothness score
among them), with \emph{synthetically injected} defects, and with episode
length controlled by \emph{truncation}.

This paper asks the natural next question: do the specific, published metrics
of \rinse{} survive these controls on real data? Our audit differs from the
Bedi testbeds on three axes, each of which closes a gap that line of work
explicitly left open:

\begin{enumerate}
\item \textbf{Named metrics, not proxies.} We implement SAL from its
published formula and TED from the \rinse{} appendix, including the
contact-boundary partition that is TED's distinguishing design feature. TED
has not previously been evaluated outside its authors' own
success-filtered setting.
\item \textbf{Real labels, not injected defects.} robomimic
can/mh~\citep{robomimic} carries \emph{human operator skill tiers}
(better/okay/worse) assigned to real teleoperators, and can/mg carries real
task-success outcomes of a scripted-policy data-generation process. No defect
is synthetic.
\item \textbf{A length-free benchmark by construction.} In can/mg every
episode is exactly 150 steps. The episode-length confound documented
by~\citep{bedi-policy} cannot exist in this data, with no truncation
required. Truncation removes the confound by discarding signal; a
fixed-horizon dataset removes it without touching the trajectories.
\end{enumerate}

The scope of the evidence is worth stating before the results rather than
after them. The detection results, which involve no training, cover two
robomimic tasks, Can and Square, and use real operator skill tiers as
labels. The success-label results cover one task, Can, because it is the
only robomimic task that ships splits containing failures. The
downstream-training results also cover Can only, under two policy
configurations: a GMM behavior-cloning policy, run for all four metrics at
ten seeds, and a diffusion policy, run for two of the four metrics at five
seeds. The downstream claims we make about TED are therefore established
under behavior cloning only, and the claim we make about jerk under the
diffusion policy holds in the configuration we tested. We make no claim
about other tasks, about human teleoperation failures, or about other
policy configurations, and Section~\ref{sec:limitations} states what each
extension would require.

Our findings are easy to state. Smoothness metrics rank \emph{operator
skill} largely to the extent that skill is confounded with how long the
episode took. Once duration is controlled, by normalization or by
duration-matched subsets, SAL's advantage collapses and raw mean jerk, the
crude baseline the smoothness literature positions itself against, becomes
the strongest detector. Against \emph{task success}, on fixed-length data,
both of \rinse{}'s metrics are at or below chance, with TED inverted, while
jerk retains signal. Curating by TED produces a behavior-cloning policy
substantially worse than random subsampling of the same size, a result that survives correction
for multiple comparisons; the weaker metrics are not distinguishable from
doing nothing. Skill and success are
different targets, and the curation literature validates on one while
deploying against the other.

\section{Related Work}

\textbf{Trajectory-statistic curation.} \rinse{}~\citep{rinse} is the primary
subject of this audit; it assumes demonstration pools pre-filtered for task
success and validates on successful trajectories only, a scoping decision its
limitations section states explicitly. PSD~\citep{psd} scores demonstrations
by total spectral power of the end-effector trajectory and reports gains over
learning-based curation on robomimic-MH and a non-expert dataset. Neither
work evaluates on pools containing failures.

\textbf{Audits of curation metrics.} Bedi's testbed
papers~\citep{bedi-audit,bedi-policy,bedi-phase} established the
detection-vs-curation decoupling, documented the episode-length confound, and
reported a negative result on per-phase metric selection. We adopt their
framing and extend it to published metrics, real labels, and a natively
fixed-length benchmark. Where they control length by truncating to a fixed
prefix, our can/mg experiments require no such intervention.

\textbf{Evaluation practice.} Kress-Gazit et
al.~\citep{empirical-science} and RoboEval~\citep{roboeval} argue that binary
success rate under-describes policy quality, and~\citep{benchmarking-audit}
shows that widely used benchmarks admit shortcut solutions. Our results are a
data-side complement: the scores used to \emph{select training data} can also
be shortcut-driven, with episode duration as the shortcut.

\section{Metrics and Datasets}
\label{sec:setup}

\subsection{What each metric measures}
\label{sec:primer}

\begin{figure}[t]
\centering
\includegraphics[width=\linewidth]{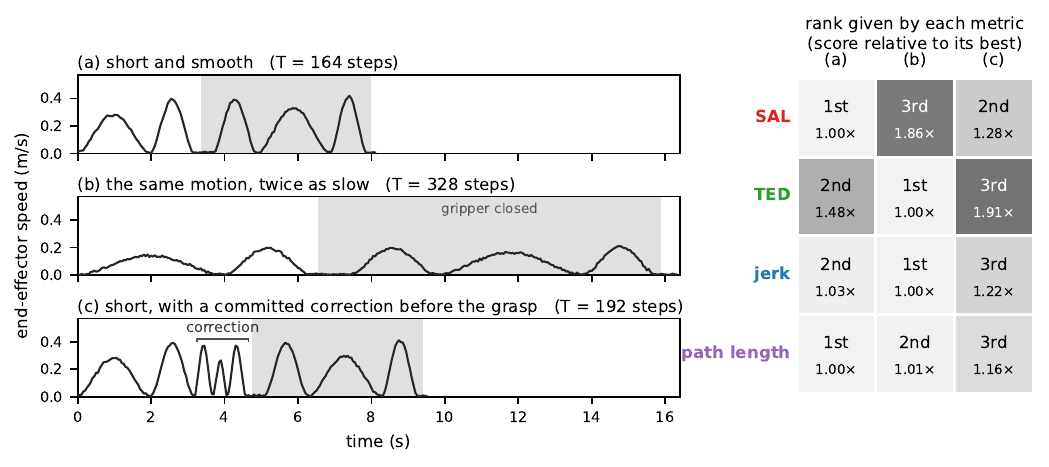}
\caption{\textbf{What the four metrics reward, on illustrative synthetic
data.} Left: end-effector speed profiles of three synthetic pick-and-place
demonstrations built from minimum-jerk segments between identical
waypoints, with the same small band-limited jitter added to each (shaded:
gripper closed). (a) is short and smooth. (b) is the same motion executed
twice as slowly, so the episode is twice as long and every segment is just
as smooth. (c) is (a) with a committed correction inserted before the
grasp: lift, shift sideways, descend again. Right: how each metric ranks
the three, computed with the implementations used throughout the paper,
with every score given relative to the best of the three under that
metric. SAL rates the slow copy (b) worst, by a wider margin than it charges
the correction in (c). TED rates the correction (c) worst and the slow copy
(b) best. Jerk also rates (c) worst and barely separates (a) from (b). Path
length cannot separate (a) from (b), which travel the same route. Every
ordering shown is stable across 30 noise seeds except the (a)/(b) near-tie
under jerk, which swaps in about a third of them, and the (a)/(c) order
under SAL, which swapped once (\texttt{scripts/fig\_metric\_primer.py}).
Synthetic and illustrative only: no number in this figure is measured on
robomimic.}
\label{fig:primer}
\end{figure}

Readers who already know these metrics can skip to
Section~\ref{sec:formal}. All four return a badness score, higher is worse,
and curating by a metric means keeping the demonstrations with the lowest
scores. The intuition behind each metric comes first here, and the formal
definitions follow in Section~\ref{sec:formal}.

\textbf{SAL (Spectral Arc Length).} SAL looks only at how fast the end
effector is moving over time, ignoring where it goes. It takes that speed
trace, breaks it into its frequency content, and measures how long and
jagged the resulting spectrum is. A speed trace made of a few clean rises
and falls, as in a single deliberate reach, has a spectrum that decays
quickly and smoothly, which gives a short arc and a good score. A trace full
of small accelerations and hesitations has content at many frequencies and
a longer arc. The metric was developed in rehabilitation science to score
the reaching movements of patients against those of healthy subjects, and
the argument for applying it to demonstrations is that a skilled operator
moves in a few committed strokes while an unskilled one makes many small
adjustments.

\textbf{TED (Trajectory-Envelope Distance).} TED asks how far the recorded
motion strays from a smoothed copy of itself. It cuts the episode into
pieces at every moment the gripper opens or closes, fits each piece with a
smooth curve (the envelope), and reports the average gap between the
recorded pose and that curve along the best time alignment of the two.
Motion that a smooth curve can follow closely scores well. Motion with
excursions the curve cannot follow scores badly. The cuts at gripper events
are what make TED contact-aware: the abrupt change of motion that
necessarily accompanies a grasp or a release falls on a segment boundary
instead of being counted as a deviation, so what remains is meant to be
the operator's wobble within each phase of the task.

\textbf{Jerk.} Jerk is the rate at which acceleration changes. Mean jerk
takes the third difference of the end-effector position at every timestep
and averages its magnitude over the episode. Motion that speeds up and
slows down gradually scores well. Abrupt starts, stops, and changes of
direction score badly. It is the classical smoothness measure in the
motor-control literature and the crude baseline that SAL and TED were
proposed to improve on. The usual criticisms are that it amplifies sensor
noise and that it does not distinguish a deliberate fast motion from a
shaky one.

\textbf{Path length.} Path length is the total distance the end effector
travels. A demonstration that goes more or less directly from start to goal
scores well. One that wanders, backtracks, or makes repeated attempts
scores badly. The intuition is that a skilled operator takes the direct
route.

\textbf{Which of these grow with episode length.} Two of the four are tied
to duration and two are not. SAL's arc is a sum over the $T/2$ frequency
bins of the spectrum, and on real data every bin carries some noise and
adds to the sum, so recording the same motion for twice as long raises SAL
even when nothing about the execution has changed. On can/mh and square/mh,
SAL's correlation with $T$ is above $+0.95$ (Section~\ref{sec:detection}).
Path length grows with distance rather than time, so a slow retrace of the
same route scores the same as a fast one, but an operator who takes longer
usually also travels further, and its correlation with $T$ is $+0.86$ on
can/paired and $+0.90$ on square/mh. Jerk and TED are averages per
timestep (TED per step of the time alignment) and do not accumulate with
$T$. TED's correlation with $T$ on can/mh is $+0.14$. Slowing the same
motion down lowers every acceleration along it, so jerk, if anything,
favors slower execution of the same path.

Figure~\ref{fig:primer} shows the two behaviors that the rest of the paper
turns on, using three synthetic speed profiles scored with our own
implementations. Demonstration (b) is demonstration (a) executed twice as
slowly, with every segment exactly as smooth, and SAL rates it the worst of
the three, by a wider margin than it charges the committed correction in
(c). SAL rewards shortness. TED rates the correction in (c) as the worst of
the three and the slow copy in (b) as the best. TED penalizes committed
motion. On real data, the first behavior appears as SAL tracking episode
length almost exactly (Section~\ref{sec:detection}), and the second as TED
rating the low-deviation hovering of failed can/mg demonstrations as
cleaner than the committed contact-phase motion of successful ones
(Section~\ref{sec:discussion}).

\subsection{Formal definitions, datasets, and statistics}
\label{sec:formal}

\textbf{Metrics.} All metrics are computed from proprioceptive signals at the
robomimic control rate of 20\,Hz ($\Delta t = 0.05$\,s) and return a badness
score (higher is worse). \emph{SAL}: spectral arc length of the end-effector
speed profile, implemented from the published formula; we negate so that
higher means less smooth. \emph{TED}: contact-aware trajectory-envelope
distance, reimplemented from the \rinse{} appendix: the 6-D pose trajectory
(position plus rotation vector, orientation weight $w_{\mathrm{ori}} =
0.25$) is partitioned at gripper open/close transitions, each segment is fit
with a greedy B\'ezier envelope, and the score is the length-normalized DTW
distance between trajectory and envelope. \emph{Jerk}: mean third-difference
magnitude of end-effector position. \emph{Path length}: total end-effector
arc length. Where indicated we also report per-timestep normalizations
(score divided by $T{-}1$). Our TED is a reimplementation from the paper, not
the authors' code; Section~\ref{sec:limitations} discusses the resulting
caveat and the sanity check we ran.

\textbf{Datasets.} All data is public robomimic~\citep{robomimic} low-dim.
\emph{can/mh}: 300 successful demonstrations from six human operators
with released skill tiers (better/okay/worse, 100 each); episode lengths
range 98--1050 steps. \emph{can/mg}: 3900 machine-generated demonstrations,
18.4\% successful, every episode exactly 150 steps. \emph{can/paired}: 200
demonstrations in matched success/failure pairs (50\% successful), lengths
55--147; we report it as a secondary setting because its matched-pair
construction makes it adversarially hard for trajectory-level scores.
\emph{square/mh}: 300 successful demonstrations with the same skill-tier
structure (100 each), episode lengths 123--1051, used in
Section~\ref{sec:square} to test whether the length confound is
task-specific. We note that Can is the only robomimic task with
failure-containing splits (\texttt{mg} and \texttt{paired}); Square,
Transport, and Tool Hang ship only proficient-human and multi-human data,
all of it successful. Our success-label experiments are therefore confined
to Can by benchmark availability, not by choice.

\textbf{Statistics.} All AUROCs are reported with 95\% bootstrap confidence
intervals (2000 resamples). Downstream comparisons use difference-of-means
bootstrap CIs (20{,}000 resamples) and Welch two-sample tests, with
Bonferroni correction applied across the full set of reported comparisons;
we mark which results survive correction and which do not. For skill we report better-vs-worse AUROC; for
success we report AUROC against the binary success label with scores negated
so that AUROC $>$ 0.5 means the metric ranks clean/successful demonstrations
higher, and AUROC $<$ 0.5 means the metric is \emph{inverted}.

\section{Detection Results}
\label{sec:detection}

\begin{figure}[t]
\centering
\includegraphics[width=\linewidth]{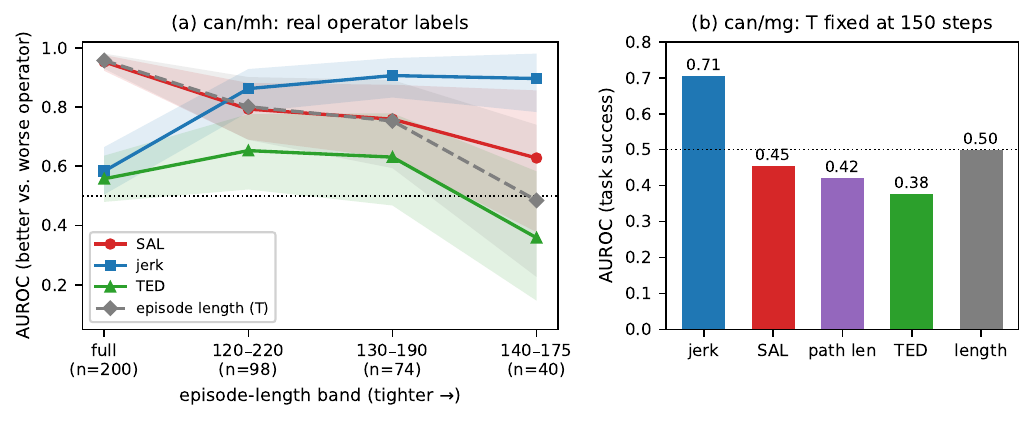}
\caption{\textbf{Smoothness metrics rank duration, not quality.}
(a) Better-vs-worse operator AUROC on can/mh across duration bands of
increasing tightness (shaded: 95\% bootstrap CI). SAL (red) tracks the
episode-length baseline (gray, dashed) almost exactly and falls from 0.952
to 0.628 as the band tightens; raw jerk (blue) moves the opposite way,
0.584 to 0.896; TED (green) never clears chance and inverts in the
tightest band. In that band episode length itself is at chance
(0.485), confirming the control. (b) On can/mg every episode is exactly
150 steps, so length carries no information by construction; against the
task-success label, jerk retains signal (0.706), SAL is at chance
(0.454), and TED is inverted (0.376).}
\label{fig:main}
\end{figure}

\subsection{Operator skill on can/mh: the length confound, on real labels}

On the full 200 better/worse demonstrations, SAL appears excellent: AUROC
0.952 [0.922, 0.975]. But episode length alone achieves 0.957 [0.929,
0.980], and SAL's correlation with $T$ is $+0.97$. The tier means make the
mechanism plain: worse operators average 304 steps, better operators 143,
and SAL's tier means (100.4 / 57.5 / 44.9) track duration almost linearly.
On this dataset SAL is, to first order, a stopwatch.

Two independent controls confirm this. First, \emph{normalization}: SAL's
arc length sums over $T/2$ spectral bins, so dividing by $T{-}1$ is the
principled per-timestep correction; normalized SAL drops from 0.952 to 0.582
[0.497, 0.662] and its length correlation drops from $+0.97$ to $+0.07$.
Second, \emph{duration matching}: restricting to bands of increasingly
similar episode length (Figure~\ref{fig:main}a, Table~\ref{tab:bands}). In
the tightest band ($T \in [140,175]$, $n{=}40$), where length itself is at
chance (0.485 [0.227, 0.740]), SAL is 0.628 [0.378, 0.856], a CI that
includes chance, while raw jerk is 0.896 [0.783, 0.980]. Across the four
bands the two metrics \emph{cross}: as the confound is removed, the crude
metric strengthens monotonically (0.584, 0.862, 0.906, 0.896) and the
sophisticated one weakens monotonically (0.952, 0.793, 0.759, 0.628).

TED never clears chance in any band (0.558 full-sample, upper CI 0.636)
and inverts to 0.359 in the tightest band. TED does \emph{not}
have the length pathology (correlation $+0.14$ with $T$); it simply carries
little skill signal on this data.

We caution that the tightest band retains only 7 worse-operator
demonstrations and its intervals are wide; the claim rests on the
\emph{trend} across all four bands and its agreement with the normalization
control, not on any single cell. We also note that na\"ive per-timestep
normalization is only appropriate for metrics whose raw value grows with
$T$; normalized path length, for example, becomes an inverse-speed proxy
that \emph{rises} with skill and is reported only for completeness in
Table~\ref{tab:bands}.

\begin{table}[t]
\centering
\small
\begin{tabular}{lcccc}
\toprule
& full ($n{=}200$) & $T \in [120,220]$ ($n{=}98$) & $T \in [130,190]$ ($n{=}74$) & $T \in [140,175]$ ($n{=}40$) \\
\midrule
SAL       & 0.952 [.922,.975] & 0.793 [.690,.881] & 0.759 [.628,.875] & 0.628 [.378,.856] \\
SAL$/T$   & 0.582 [.497,.662] & 0.600 [.479,.715] & 0.603 [.446,.755] & 0.658 [.342,.923] \\
TED       & 0.558 [.480,.636] & 0.653 [.522,.776] & 0.631 [.468,.779] & 0.359 [.147,.583] \\
jerk      & 0.584 [.505,.665] & 0.862 [.783,.928] & 0.906 [.832,.965] & 0.896 [.783,.980] \\
path len. & 0.783 [.718,.842] & 0.590 [.449,.730] & 0.624 [.461,.781] & 0.429 [.215,.645] \\
\midrule
length $T$ & 0.957 [.929,.980] & 0.801 [.689,.901] & 0.753 [.594,.891] & \textbf{0.485} [.227,.740] \\
\bottomrule
\end{tabular}
\caption{Better-vs-worse operator AUROC on can/mh across duration bands,
with 95\% bootstrap CIs. The bottom row is the control: a band is
duration-matched only where $T$ itself is near chance. Per-timestep
normalizations of TED and path length are omitted: unlike SAL, whose raw
value grows with the number of spectral bins, these quantities do not scale
linearly with $T$, and dividing by $T$ turns them into inverse-speed
proxies rather than duration-free measures of smoothness.}
\label{tab:bands}
\end{table}

\subsection{Task success on can/mg: the confound removed by construction}

The can/mg split fixes every episode at 150 steps, so no metric can exploit
duration. Against the success label (716 of 3900 successful), jerk achieves
0.706, SAL 0.454, path length 0.421, and TED 0.376
(Figure~\ref{fig:main}b). The length baseline is exactly 0.500, as it must
be. Per-timestep normalization leaves every AUROC unchanged, as expected
under a constant divisor, which we report as a consistency check.

Two readings of this result matter. First, both of \rinse{}'s metrics are at
or below chance on the label practitioners actually deploy against. TED's
inversion means that filtering by TED \emph{preferentially retains failed
demonstrations}. Second, the ordering among metrics reverses relative to the
uncontrolled skill setting: the metric that looked worst there (jerk) is the
only one with signal here.

\subsection{Secondary setting: can/paired}

On the matched-pair dataset every metric is inverted against success (TED
0.331, SAL 0.398, jerk 0.368, path length 0.029), but the setting is
confounded in the opposite direction: successful demonstrations are
systematically \emph{longer} (length-alone AUROC 0.194), and path length's
$+0.86$ correlation with $T$ makes it a pure length detector here. We
report can/paired for completeness and treat can/mg as the primary
success-label evidence.

\subsection{Generalization to a second task: square/mh}
\label{sec:square}

If the length confound were an artifact of the Can task, the argument would
be considerably weaker. We therefore repeat the skill-label analysis on
square/mh, a harder insertion task with the same operator skill-tier
structure and a comparable duration range (123--1051 steps versus Can's
98--1050). No retuning was performed: the same metric implementations and
the same band procedure were applied to the new data.

The confound replicates almost exactly. SAL's correlation with episode
duration is $+0.96$ on Square against $+0.97$ on Can, and its uncontrolled
skill AUROC is 0.898 [0.852, 0.936] against a length-alone baseline of 0.936
[0.902, 0.964]. Per-timestep normalization collapses it to 0.458 [0.376,
0.536], slightly below chance and a larger drop than the corresponding Can
figure. Path length behaves the same way, reaching 0.931 [0.893, 0.963]
uncontrolled with a $+0.90$ duration correlation.

The crossover also replicates (Table~\ref{tab:square}). As the duration band
tightens and the length baseline falls from 0.936 to 0.407, SAL falls
monotonically (0.898, 0.759, 0.529, 0.480) while jerk rises monotonically
(0.431, 0.570, 0.782, 0.896). In the band where duration is genuinely
controlled ($T \in [200,240]$, length AUROC 0.407), jerk reaches 0.896
[0.724, 1.000], numerically identical to its value in Can's tightest band,
while SAL sits at 0.480 [0.250, 0.701], a CI centred on chance.

Two differences from Can are worth stating. First, jerk is \emph{inverted}
on Square before any control is applied (0.431), where on Can it was weakly
positive (0.584); the uncontrolled ranking of metrics is therefore itself
task-dependent, even though the controlled ranking is not. Second, path
length remains strong in the controlled band on Square (0.887 [0.742,
0.991]) where it collapsed on Can (0.429 [0.215, 0.645]). We do not claim
jerk or path length is a good curation metric on the strength of these
numbers; the point is narrower, that whatever advantage SAL displays over
them disappears once duration is held fixed, on both tasks. The tightest
Square band retains 13 worse-operator demonstrations, so as with Can the
claim rests on the monotone trend across bands rather than on any single
cell.

\begin{table}[t]
\centering
\small
\begin{tabular}{lcccc}
\toprule
& full ($n{=}200$) & $T \in [150,300]$ ($n{=}111$) & $T \in [180,260]$ ($n{=}55$) & $T \in [200,240]$ ($n{=}30$) \\
\midrule
SAL       & 0.898 [.852,.936] & 0.759 [.658,.849] & 0.529 [.344,.702] & 0.480 [.250,.701] \\
SAL$/T$   & 0.458 [.376,.536] & 0.485 [.372,.602] & 0.468 [.300,.653] & 0.520 [.280,.738] \\
TED       & 0.663 [.584,.736] & 0.685 [.577,.788] & 0.599 [.427,.758] & 0.557 [.335,.785] \\
jerk      & 0.431 [.349,.509] & 0.570 [.455,.687] & 0.782 [.618,.923] & 0.896 [.724,1.000] \\
path len. & 0.931 [.893,.963] & 0.817 [.726,.895] & 0.796 [.654,.913] & 0.887 [.742,.991] \\
\midrule
length $T$ & 0.936 [.902,.964] & 0.824 [.738,.898] & 0.613 [.456,.766] & \textbf{0.407} [.197,.629] \\
\bottomrule
\end{tabular}
\caption{Better-vs-worse operator AUROC on \emph{square}/mh across duration
bands, with 95\% bootstrap CIs, in the same format as
Table~\ref{tab:bands}. The bottom row is the control. The SAL/jerk
crossover observed on Can reproduces here: SAL tracks the length baseline
down while jerk rises, and in the band where length is at chance jerk
reaches the same 0.896 it reaches in Can's tightest band.}
\label{tab:square}
\end{table}

\section{Does Detection Predict Curation? Downstream Training}
\label{sec:training}

Detection AUROC is not the quantity practitioners care
about~\citep{bedi-policy}; the trained policy is. We therefore filter can/mg
by each metric at a 50\% keep fraction and train a GMM behavior-cloning
policy (MLP $1024{\times}1024$, 5 modes, batch 100, 600 epochs of 500
gradient steps, learning rate $10^{-4}$) on each curated subset, with 50
evaluation rollouts (horizon 400) every 150 epochs and the best checkpoint
reported per seed. Conditions: TED, SAL, jerk, path length, plus three
references: an \emph{oracle} keeping only successful demonstrations, a
\emph{random} 50\% subset, and an uncurated size-matched control. Ten seeds
per condition. All filtered pools contain 1950 demonstrations except the
oracle (716, the number of successes that exist); we therefore treat the
oracle as a ceiling reference rather than a matched condition.

\begin{table}[t]
\centering
\small
\begin{tabular}{lccc}
\toprule
condition & pool size & best success rate (mean $\pm$ std) & 95\% CI \\
\midrule
oracle (successes only) & 716 & 0.892 $\pm$ 0.048 & [0.866, 0.922] \\
\midrule
no curation             & 1950 & 0.380 $\pm$ 0.063 & [0.344, 0.420] \\
random 50\%             & 1950 & 0.360 $\pm$ 0.061 & [0.326, 0.398] \\
\midrule
SAL                     & 1950 & 0.332 $\pm$ 0.061 & [0.298, 0.370] \\
jerk                    & 1950 & 0.332 $\pm$ 0.041 & [0.306, 0.354] \\
path length             & 1950 & 0.310 $\pm$ 0.030 & [0.292, 0.328] \\
TED                     & 1950 & 0.240 $\pm$ 0.045 & [0.212, 0.266] \\
\bottomrule
\end{tabular}
\caption{Downstream GMM-BC policy success on can/mg after filtering to a
fixed 1950-demonstration pool by each metric, 10 seeds each. The oracle pool
is smaller (716, the number of successes that exist) and is a ceiling
reference, not a matched condition. Pairwise tests against the baselines are
in Table~\ref{tab:tests}.}
\label{tab:training}
\end{table}

\begin{table}[t]
\centering
\small
\begin{tabular}{lcccc}
\toprule
& \multicolumn{2}{c}{vs.\ random 50\%} & \multicolumn{2}{c}{vs.\ no curation} \\
\cmidrule(lr){2-3}\cmidrule(l){4-5}
metric & $\Delta$ [95\% CI] & $p$ & $\Delta$ [95\% CI] & $p$ \\
\midrule
TED         & $-0.120$ [$-0.166$, $-0.076$] & $\mathbf{0.0001}$ & $-0.140$ [$-0.186$, $-0.096$] & $\mathbf{<0.0001}$ \\
path length & $-0.050$ [$-0.090$, $-0.012$] & 0.037 & $-0.070$ [$-0.112$, $-0.030$] & 0.0071 \\
jerk        & $-0.028$ [$-0.072$, $+0.014$] & 0.25 & $-0.048$ [$-0.092$, $-0.006$] & 0.060 \\
SAL         & $-0.028$ [$-0.080$, $+0.024$] & 0.32 & $-0.048$ [$-0.098$, $+0.002$] & 0.098 \\
\bottomrule
\end{tabular}
\caption{Difference in mean best success rate between each metric-curated
condition and the two baselines, with 95\% bootstrap CIs on the difference
(20{,}000 resamples) and Welch two-sample $p$-values. Bold marks the
comparisons surviving Bonferroni correction across all eight tests
($\alpha = 0.05/8 = 0.006$). Only TED clears that bar. For reference,
random and no-curation are themselves indistinguishable
($\Delta = -0.020$, CI $[-0.072, +0.032]$).}
\label{tab:tests}
\end{table}

Table~\ref{tab:training} answers the question Section~\ref{sec:detection}
could only pose, and Table~\ref{tab:tests} reports the comparisons that
matter. Two results are unambiguous. Curating by TED costs roughly a third
of downstream success relative to random selection of the same size
($-0.120$, $p = 0.0001$) and relative to not curating at all ($-0.140$,
$p < 0.0001$); both survive Bonferroni correction across the eight
comparisons we report. Path length is directionally harmful with weaker
support ($p = 0.037$ and $p = 0.0071$), clearing conventional significance
but not correction. SAL and jerk sit below both baselines in the mean but
are not statistically distinguishable from them ($p \geq 0.06$), and we
report them as such rather than as evidence of harm.

We emphasize this because an earlier version of this analysis, run with
5 seeds rather than 10, appeared to show every metric-curated condition
underperforming both baselines. That pattern did not survive the seed
expansion for SAL and jerk. Since the argument of this paper is that
under-powered evaluation produces confident claims in the wrong direction,
we apply the same standard to our own results: the defensible claim is about
TED, not about smoothness metrics as a class.

What the grid does establish irrespective of any single comparison is that
under this policy configuration none of these metrics recovers usable signal that is
demonstrably present.
An oracle keeping only true successes reaches 0.892 [0.866, 0.922], roughly
2.5$\times$ the best metric-curated condition, so the pool contains ample
information about which demonstrations to keep; the trajectory statistics
simply do not find it. Random subsampling and the uncurated control are
themselves indistinguishable ($\Delta = -0.020$, CI spanning zero), which is
the sanity check one wants: at a 50\% keep fraction on this data, discarding
demonstrations at random costs nothing measurable, so any deficit a metric
shows is attributable to \emph{which} demonstrations it discards rather than
to the reduction in pool size.

The downstream ordering tracks the detection ordering imperfectly. TED is
worst in both, and path length is weak in both, but jerk has the best
detection AUROC on can/mg (0.706) while landing indistinguishable from SAL
downstream, whose detection AUROC is near chance (0.454). Detection accuracy
is at best a coarse predictor of curation value, consistent with the
decoupling reported by~\citep{bedi-policy} on a different testbed.

\subsection{Cross-architecture replication under a diffusion policy}
\label{sec:crossarch}

The grid above uses a single policy configuration, so on its own it cannot
distinguish a property of the metric from a property of behavior cloning. We
therefore replicated it under a diffusion policy on the same curated
subsets: same keep fraction, same 1950-demonstration pools, same 600 epochs
of 500 gradient steps, same learning rate ($10^{-4}$), and the same
evaluation protocol (50 rollouts, horizon 400, every 150 epochs, best
checkpoint per seed).\footnote{Reporting the best checkpoint per seed is a
maximum over the eight evaluations each run receives, and is therefore
biased upward; both grids are evaluated eight times, so the bias is
symmetric across architectures. Recomputing every comparison with the
\emph{mean} over the eight evaluations instead leaves all conclusions
intact, with one exception we record here: TED versus no curation under
diffusion reaches $p = 0.0110$ under the mean estimator and would survive
this architecture's correction threshold ($\alpha = 0.0125$), whereas under
the reported best-of-eight estimator it does not ($p = 0.11$). TED versus
random remains non-significant under both ($p = 0.12$ and $p = 0.17$). We
report the conservative version throughout.} The policy is a conditional
U-Net ($[256, 512, 1024]$
down-dims, kernel 5, 8 groups, 256-dimensional diffusion-step embedding)
with EMA (power 0.75), an observation horizon of 2, action horizon of 8, and
prediction horizon of 16, using a DDIM noise scheduler with 10 inference
steps. Batch size (256 vs.\ 100) and the sequence modeling are inherent to
the architecture and are the only respects in which the two grids differ by
more than the policy configuration. We ran five conditions (TED, jerk, and the three
references) with five seeds each; SAL and path length were not rerun, so the
diffusion comparison covers two of the four metrics.

\begin{table}[t]
\centering
\small
\begin{tabular}{lcc}
\toprule
condition & GMM-BC (10 seeds) & diffusion (5 seeds) \\
\midrule
oracle (successes only) & 0.892 $\pm$ 0.048 & 0.544 $\pm$ 0.030 \\
\midrule
no curation             & 0.380 $\pm$ 0.063 & 0.120 $\pm$ 0.014 \\
random 50\%             & 0.360 $\pm$ 0.061 & 0.116 $\pm$ 0.017 \\
\midrule
jerk                    & 0.332 $\pm$ 0.041 & 0.212 $\pm$ 0.030 \\
TED                     & 0.240 $\pm$ 0.045 & 0.092 $\pm$ 0.030 \\
\bottomrule
\end{tabular}
\caption{Mean best success rate under both policy configurations, restricted to the
five conditions run under both. TED is the worst condition under each. Jerk
moves from indistinguishable-from-baseline under BC to above both baselines
under diffusion. Absolute rates are lower under diffusion in every
condition, by factors ranging from $1.6\times$ (oracle) to $3.1\times$ (the
two baselines).}
\label{tab:crossarch}
\end{table}

\begin{table}[t]
\centering
\small
\begin{tabular}{lcccc}
\toprule
& \multicolumn{2}{c}{vs.\ random 50\%} & \multicolumn{2}{c}{vs.\ no curation} \\
\cmidrule(lr){2-3}\cmidrule(l){4-5}
metric & $\Delta$ [95\% CI] & $p$ & $\Delta$ [95\% CI] & $p$ \\
\midrule
jerk & $+0.096$ [$+0.068$, $+0.124$] & $\mathbf{0.0007}$ & $+0.092$ [$+0.064$, $+0.116$] & $\mathbf{0.0011}$ \\
TED  & $-0.024$ [$-0.048$, $+0.004$] & 0.17 & $-0.028$ [$-0.052$, $+0.000$] & 0.11 \\
\bottomrule
\end{tabular}
\caption{Diffusion-policy grid, same procedure as Table~\ref{tab:tests}:
95\% bootstrap CIs on the difference (20{,}000 resamples) and Welch
two-sample $p$-values. Bold marks comparisons surviving Bonferroni
correction across the four tests reported for this architecture
($\alpha = 0.05/4 = 0.0125$). Random and no-curation are again
indistinguishable ($\Delta = -0.004$, CI $[-0.020, +0.012]$).}
\label{tab:crossarch_tests}
\end{table}

Three results follow, and they do not all point the same way.

First, TED is the worst condition under both architectures (0.240 under BC,
0.092 under diffusion). That is what one expects if the deficit belongs to
the metric rather than to behavior cloning, and the replication is
consistent with that part of the original finding. But the harm is statistically
robust under BC only. Under diffusion the deficit is $-0.024$ against random
and $-0.028$ against no curation, with intervals that admit zero, and
neither survives correction. This is a statement about power rather than a
refutation: at five seeds this grid resolves differences of roughly 0.054,
essentially the same threshold we report for the BC grid, and TED's deficit
falls below it. The defensible claim is that TED's \emph{ordering} holds
across architectures while its \emph{harm} is demonstrated under BC only.
Descriptively, and not as a tested comparison, the deficit is of broadly
similar relative size in the two grids: 37\% of the no-curation baseline
under BC and 23\% under diffusion. We report this to characterize how the
compressed diffusion success scale relates to the absolute differences in
Table~\ref{tab:crossarch_tests}, not as evidence that the two effects are
equal.

Second, jerk reverses sign. Under BC it is indistinguishable from both
baselines ($p \geq 0.06$); under diffusion it beats both by $+0.096$ and
$+0.092$, and both comparisons survive correction. This is the largest
effect in the diffusion grid, and it runs opposite to the paper's headline:
at least one crude trajectory statistic, never proposed as a curation
metric, has genuine curation value in the configuration we tested. It is
also the second time in this work that jerk outperforms the metrics
motivated against it, the first being detection inside a duration-matched
band. What may drive the reversal is taken up below.

Third, absolute success rates are lower under diffusion in every condition,
by factors ranging from $1.6\times$ on the oracle (0.544 vs.\ 0.892) to
$3.1\times$ on the two baselines, at a matched epoch and evaluation budget
in which the diffusion runs in fact saw more samples per epoch (batch 256
vs.\ 100). The gap is therefore not a uniform offset: it is widest exactly
on the mixed-quality pools the curation comparison runs on.

One explanation can be ruled out. The diffusion runs show no measurable
improvement over the second half of training: averaged across all 25 runs,
mean success in the last four evaluations exceeds the first four by
$+0.017$, against a binomial noise floor of $0.044$ for 50-rollout
evaluations, and only 3 of 25 runs improve by more than one noise standard
deviation. Per-run variation across the eight evaluations has a standard
deviation $1.01\times$ the noise floor, which is to say it is
indistinguishable from sampling noise alone. The identical measurement on
the BC grid gives $+0.092$, with 56 of 70 runs improving and a ratio of
$1.46\times$, so the flatness is a property of the diffusion runs and not
of our evaluation protocol. Whatever limits these policies is therefore not
training duration; they converge early and stop. We record this because
under-training is the most natural explanation for the gap and it does not
survive contact with the training curves, which narrows the space of
remaining explanations without confirming one. One candidate is action
parameterization: our runs used robomimic's default delta actions, which
Chi et al.~\citep{chi-dp} report handicap diffusion policy while helping BC,
so the gap and the jerk reversal may partly reflect it, and the policy-class
reading of both is provisional. The consequence for reading
Table~\ref{tab:crossarch} is concrete: the two
columns are not operating at comparable points on the success scale, so
cross-architecture comparisons of absolute differences are weaker than
within-column comparisons, and the claims above are deliberately about
ordering and sign rather than magnitude.

We correct within each architecture, since each grid is a separate family of
comparisons. Pooling all twelve tests instead ($\alpha = 0.05/12 = 0.00417$)
changes nothing: TED's two BC comparisons and jerk's two diffusion
comparisons still survive, and path length's $p = 0.0071$ still does not.

Taken together, the replication is consistent with the metric-level reading of the TED
result and simultaneously narrows the scope of any claim about smoothness
metrics as a class. Which demonstrations are worth keeping is not a property
of the demonstrations alone, and a curation result reported without naming
the policy configuration it was measured on, architecture and action
parameterization included, omits something that can change the sign of the
answer.

\section{Discussion}
\label{sec:discussion}

\begin{figure}[t]
\centering
\includegraphics[width=\linewidth]{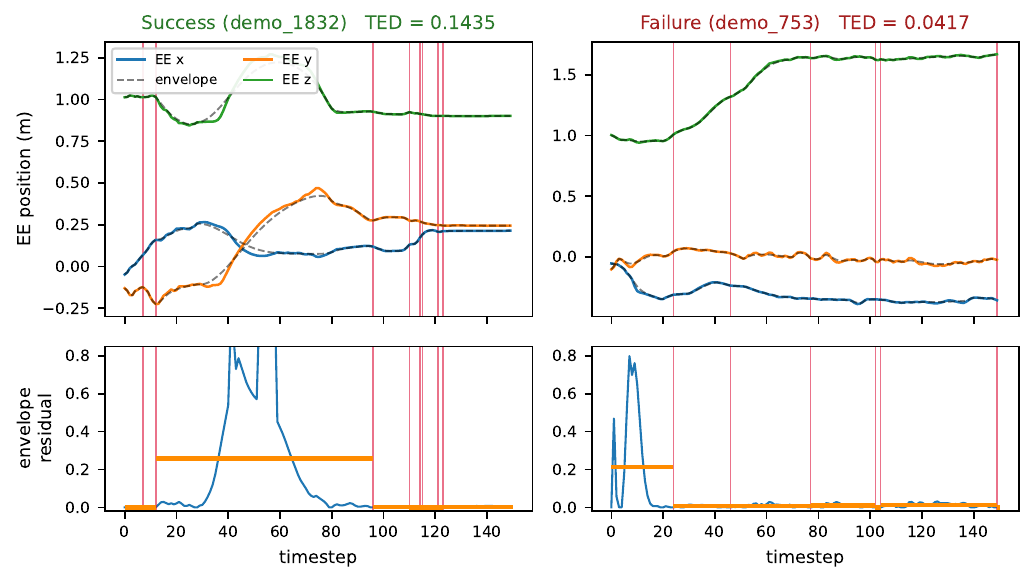}
\caption{\textbf{Failure stays inside the envelope; success does not.} A successful (left)
and failed (right) can/mg demonstration, with end-effector position, TED's
gripper-contact segment boundaries (red), the per-segment B\'ezier envelope
(dashed), and the per-timestep envelope residual (bottom, orange bars are
per-segment means). Both trajectories contain one high-deviation segment,
but in the success it spans timesteps 12--96 and coincides with the
descend-and-grasp, while in the failure it is confined to the initial
free-space approach (0--24) and the remaining 126 steps are a flat tail of
repeated gripper cycles that never completes a placement. TED averages over
the alignment path and so rates the failure as the cleaner trajectory.
Illustrative of the mechanism only; the quantitative claim is
Figure~\ref{fig:main}b.}
\label{fig:qual}
\end{figure}

\textbf{Skill and success are different targets.} The same metric family
gives opposite verdicts depending on the label. Under an uncontrolled skill
label, SAL looks nearly perfect; under a length-controlled skill label it is
marginal; under a success label on fixed-length data it is at chance and TED
is inverted. Curation papers validate against operator skill or
smoothness-adjacent labels on success-only pools, then are deployed, in
practice, to remove \emph{failures} from mixed pools. Our results say those
are not the same job, and the metrics do not transfer between them.

\textbf{The crude baseline wins under controls.} \rinse{} motivates SAL and
TED as principled improvements over na\"ive statistics such as jerk. Under
every length control we applied, that ordering reverses. We do not claim
jerk is a good curation metric in general, and the Bedi
testbeds~\citep{bedi-audit} show smoothness-family scorers failing badly on
structural defects; we claim only that on real robomimic data the published
sophistication does not survive the controls that the audit literature has
shown to be necessary.

\textbf{Why TED inverts: failure stays inside the envelope.} TED's design premise is that
partitioning at gripper-contact boundaries isolates task-relevant
abruptness, and a plausible hypothesis going in was that this would let TED
catch failures, since canonical structural failures are contact-boundary
events. The data supports a different and simpler mechanism.
Figure~\ref{fig:qual} contrasts a successful and a failed can/mg
demonstration. Both contain one high-deviation segment, but they differ in
when it occurs and how much of the episode it spans. In the success, the
high-residual segment runs from timestep 12 to 96 (84 of 150 steps, mean
residual 0.259) and coincides with the descend-and-grasp: the end effector
dips, the lateral axes swing, and the envelope cannot track the motion. In
the failure, the only comparable deviation is the initial free-space
approach (timesteps 0--24, mean residual 0.213); after that the trajectory
settles into a flat tail in which the end effector holds height while the
gripper opens and closes repeatedly without completing a placement, and
every remaining segment has residual below 0.014. Because TED averages
deviation over the DTW alignment path, a committed motion spanning half the
episode raises the score substantially while a short approach transient is
diluted by the smooth remainder. TED therefore rates the failure as the
cleaner trajectory (0.042 vs.\ 0.144).

On this task, in other words, successful manipulation \emph{requires}
sustained high-deviation motion through the contact phase, while failure
manifests as low-amplitude hovering that the envelope follows closely. A
metric that rewards agreement with a smoothed envelope is measuring
commitment with the sign flipped. Jerk sees the same two behaviors from
the other side, which is why it is the one metric that retains signal in
Section~\ref{sec:detection}. A committed motion is large but smooth: it
carries the end effector far from the envelope, but its accelerations
change gradually, so its envelope deviation is high and its jerk is low.
Hovering is small but jittery: the end effector holds position through
repeated gripper cycles, and what little it moves, it moves abruptly, so
its envelope deviation is low and its jerk is high. The two demonstrations in Figure~\ref{fig:qual} show
both readings at once, with the success at a mean jerk of 71.6 against
124.1 for the failure while TED rates the failure cleaner, and across the
whole split successful demonstrations have lower mean jerk than failed
ones (85.7 vs.\ 112.8), which is the direction the 0.706 AUROC reports.
We stress that these panels illustrate a mechanism rather than
establish it; the quantitative claims are the AUROCs of 0.376 and 0.706 in
Figure~\ref{fig:main}b, and the per-demonstration TED scores we inspected
(successes 0.144 and 0.079, failures 0.042, 0.042, and 0.017) are
consistent with but not evidence for those numbers.

\section{Limitations}
\label{sec:limitations}

Our detection results cover two robomimic tasks (Can and Square) but our
success-label and downstream-training results cover only Can, because Can is
the only robomimic task shipping failure-containing splits; extending the
success-label analysis requires datasets with real, labeled teleoperation
failures, such as \mbox{REASSEMBLE}~\citep{reassemble} or
DROID~\citep{droid}. Our downstream results cover two policy configurations (GMM behavior
cloning and a diffusion policy); the detection results are policy-free. The
cross-architecture grid of Section~\ref{sec:crossarch} carries three
limitations of its own. It covers two of the four metrics, since SAL and
path length were not rerun under diffusion, so the claim that TED ranks
worst under both architectures rests on a five-condition subset on the
diffusion side against a seven-condition grid on the BC side. It uses five
seeds rather than ten, resolving differences of roughly 0.054; TED's
diffusion deficit ($-0.028$, CI $[-0.052, +0.000]$) falls below that, which
is why we report its ordering as replicating and its harm as established
under BC only. And absolute success rates remain lower under diffusion in
every condition at a matched epoch and evaluation budget, from $1.6\times$
on the oracle (0.544 vs.\ 0.892) to $3.1\times$ on the two baselines. We
previously attributed that gap to under-convergence in a reduced-compute
pilot; restoring the full epoch and rollout budget did not close it, and the
training curves rule that explanation out directly, since the diffusion runs
improve by only $+0.017$ between the first and second halves of training
against a $0.044$ noise floor while the BC runs improve by $+0.092$. The gap
is therefore not a matter of training duration; action parameterization
(Section~\ref{sec:crossarch}) is a candidate we have not yet tested. Because the two grids sit at different points on the
success scale, we restrict cross-architecture claims to ordering and sign
rather than magnitude. Multiple-comparison correction is applied within each
architecture, as each grid is a separate family; pooling all twelve tests
($\alpha = 0.00417$) leaves every reported conclusion unchanged.
Our TED is a reimplementation from the \rinse{} appendix, not the authors'
released code; as a sanity check it broadly agrees with SAL's ranking on the
clean proficient-human data, but a discrepancy with the reference
implementation cannot be excluded, and we will replace it if code becomes
available. The tightest duration band retains only 7 negatives, and its
role in our argument is corroborative, not primary. The BC downstream
conditions use 10 seeds; at this power, differences smaller than roughly
0.05 in mean success rate are not resolvable, which is why we report SAL and
jerk as indistinguishable from the baselines rather than as harmful under
that architecture. The oracle training
condition has a smaller pool (716 vs.\ 1950) and is reported as a ceiling,
not a matched comparison. Finally, our success-label experiments use
machine-generated failures (can/mg); failure modes of human teleoperators
may present differently. We report can/paired only as a secondary setting
because its matched-pair construction, in which each success/failure pair
shares a near-identical trajectory prefix and diverges only in the final
placement, concentrates the discriminating signal into a handful of
terminal timesteps and simultaneously introduces an opposing length
confound (successful demonstrations are systematically longer, length-alone
AUROC 0.194); neither property is representative of the mixed-quality pools
curation metrics are deployed on.

\section{Conclusion}

Published trajectory-smoothness curation metrics, evaluated exactly where
their authors did not, on real operator labels under duration controls and
on fixed-length data against task success, rank episode duration more than
demonstration quality. Once duration is controlled, the crude jerk baseline
they were designed to supersede outperforms them, and the contact-aware
metric inverts, and the same crossover reproduces on a second task without
retuning. Training on the metrics' own curated subsets shows the practical
stakes are real but narrower than a five-seed pass suggested: curating by
TED costs roughly a third of downstream success against random selection of
the same size, a result robust to correction for multiple comparisons, while
the other metrics are directionally below the baselines without being
distinguishable from them. An oracle reaches 0.892 on the same pool, so the
signal these metrics fail to find is unambiguously present in the data.
In the diffusion configuration we tested, TED still ranks worst under both
architectures, consistent with a metric-level rather than architecture-level
reading of that result, though at five seeds its deficit there is not
resolvable and the harm remains established under behavior cloning only. The
same replication reverses jerk from inert to better than both baselines. A
curation metric's value, and its sign, depend on the policy configuration it
is measured against, which is not currently part of how such results are
reported.

\bibliographystyle{plainnat}

\end{document}